%% file: ms.tex
\documentclass[conference]{IEEEtran}
\IEEEoverridecommandlockouts
\usepackage{cite}
\usepackage{amsmath,amssymb,amsfonts}
\usepackage{algorithmic}
\usepackage{graphicx}
\usepackage{textcomp}
\usepackage{xcolor}
\usepackage{subcaption}
\usepackage{adjustbox}
\usepackage{multicol}
\usepackage{multirow}
\usepackage{flushend}
\usepackage{url}
\usepackage{booktabs}

\usepackage[pagebackref,breaklinks,colorlinks]{hyperref}

\newlength\savewidth

\def\BibTeX{{\rm B\kern-.05em{\sc i\kern-.025em b}\kern-.08em
    T\kern-.1667em\lower.7ex\hbox{E}\kern-.125emX}}
\input{def.set}

\newcommand{\authorskip}{\hspace{2.8mm}}
\newcommand{\emailskip}{\hspace{4mm}}

\begin{document}

\title{Predicting Adverse Neonatal Outcomes for Preterm Neonates with Multi-Task Learning}

\author{
 Jingyang Lin$^{1}$ \authorskip Junyu Chen$^{1}$ \authorskip Hanjia Lyu$^{1}$ \authorskip
 Igor Khodak$^{2}$ \authorskip Divya Chhabra$^{2}$  \\ \authorskip Colby L Day Richardson$^{2}$ \authorskip Irina Prelipcean$^{2}$  \authorskip Andrew M Dylag$^{2}$ \authorskip Jiebo Luo$^{1}$ \\[2mm]
 $^{1}$University of Rochester  \qquad $^{2}$University of Rochester Medical Center\\
 \texttt{\small \{jlin81,jchen175,hlyu5\}@ur.rochester.edu} \emailskip \texttt{\small jluo@cs.rochester.edu}\\[0mm]
 \texttt{\small \{igor\_khodak,divya\_chhabra,colby\_richardson,irina\_prelipcean,andrew\_dylag\}@urmc.rochester.edu}
}

\maketitle

\input{text/abstract}

\input{text/introduction}

\input{text/relatedwork}

\input{text/data}

\input{text/method}

\input{text/experiment}

\input{text/conclusion}

{\small
\bibliographystyle{ieee_fullname}
\bibliography{egbib}
}

\end{document}

%% file: text/abstract.tex
\begin{abstract}
Diagnosis of adverse neonatal outcomes is crucial for preterm survival since it enables doctors to provide timely treatment. Machine learning (ML) algorithms have been demonstrated to be effective in predicting adverse neonatal outcomes. However, most previous ML-based methods have only focused on predicting a single outcome, ignoring the potential correlations between different outcomes, and potentially leading to suboptimal results and overfitting issues. In this work, we first analyze the correlations between three adverse neonatal outcomes and then formulate the diagnosis of multiple neonatal outcomes as a multi-task learning (MTL) problem. We then propose an MTL framework to jointly predict multiple adverse neonatal outcomes. In particular, the MTL framework contains shared hidden layers and multiple task-specific branches. Extensive experiments have been conducted using Electronic Health Records (EHRs) from 121 preterm neonates. Empirical results demonstrate the effectiveness of the MTL framework. Furthermore, the feature importance is analyzed for each neonatal outcome, providing insights into model interpretability.
\end{abstract}


%% file: text/introduction.tex
\section{Introduction}
\label{sec:intro}

The neonatal period, encompassing the initial four weeks of an infant's life, marks a crucial stage of development characterized by significant physiological changes and high vulnerability that leads to an elevated neonatal mortality rate~\cite{karimi2019mortality}. Around two-thirds of infant deaths occur during this period~\cite{daemi2019risk}. Predictions indicate that approximately 26 million newborns will die between 2019 and 2030~\cite{rezaeian2020prediction}, making the issue of high neonatal mortality a global concern. Despite the remarkable advancements in neonatal care, infants born before 37 weeks of pregnancy still face a substantial risk of mortality~\cite{richter2019temporal}. In 2021, one-tenth of babies were born too early in the United States.\footnote{\scriptsize\url{https://www.cdc.gov/reproductivehealth/features/premature-birth}} As a result, it is imperative that we prioritize our attention towards Extremely Low Gestational Age Newborns (ELGANs, i.e., less than 28 weeks of gestation) rather than newborns in general.

\textit{Adverse neonatal outcomes} play a significant role in neonatal survival~\cite{workineh2022adverse}. Early diagnosis of adverse neonatal outcomes enables physicians to prepare an early treatment. It also allows researchers to identify the risk factors that lead to a high neonatal mortality rate~\cite{houweling2019prediction}.

\input{figures/data_stat}

Machine learning (ML) is a branch of artificial intelligence. Given a model and a set of labeled/unlabeled data, ML methods learn to recognize informative patterns from data without explicit programming~\cite{jordan2015machine}. The ML model can make reasonable predictions on unseen test data based on these learned patterns~\cite{al2019survey}. A series of innovations~\cite{mboya2020prediction,sheikhtaheri2021prediction,nguyen2021whom,hsu2021machine,mangold2021machine,turova2020machine,hsu2021machine2} have been proposed to predict adverse neonatal outcomes using machine learning techniques. However, most previous works consider the prediction of various neonatal outcomes as multiple \textit{single-task} learning problems. These methods \textbf{ignore the correlation among different outcomes}, which leads to \textbf{inefficient usage of data}~\cite{standley2020tasks}. Moreover, single-task learning is prone to \textbf{overfitting}~\cite{ruder2017overview,zhang2021survey}, especially when the amount of annotated data is limited, which is often the case with many medical applications. Therefore, this work aims to \textbf{diagnose multiple neonatal outcomes jointly  to improve data efficiency} by leveraging the correlations among these outcomes.

In this work,  we first collect a dataset including 121 preterm neonates from two medical centers and focus on three adverse neonatal outcomes, including severe bronchopulmonary dysplasia (BPD), pulmonary hypertension (PH) diagnosis, and discharge weight. In particular, BPD~\cite{jobe2001bronchopulmonary} is a form of chronic lung disease, which most often occurs in low-weight neonates born more than two months early. Pulmonary hypertension~\cite{hoeper2013definitions} is when the blood pressure in the lungs' arteries is elevated. The discharge weight is the preterm neonates' weight when discharged from the hospital after birth. Next, we analyze the data distribution and correlations between these outcomes. The results show that the three adverse neonatal outcomes are relevant to each other. For instance, infants with more weight gain might have less risk of BPD and PH. Therefore, we propose to make joint predictions of multiple adverse neonatal outcomes for preterm neonates to improve the accuracy of individual outcome predictions.

Moreover, by considering the correlations across different neonatal outcomes, we first formulate the diagnosis of multiple neonatal outcomes as a multi-task learning (MTL) problem. We then propose an MTL framework to solve this problem. Technically, our proposed multi-task learning framework is based on a Neural Network (NN) model. It consists of shared hidden layers and multiple task-specific hidden layers followed by predictors for different tasks. In particular, the shared hidden layers aim to capture \textbf{correlated yet often hidden knowledge among all neonatal outcomes}. Different task-specific branches are designed to learn the \textbf{unique features of each neonatal outcome}. By leveraging the correlations and task-specific information of the outcomes, this framework pursues the joint learning of multiple adverse neonatal outcomes. In addition, it enhances data efficiency and mitigates the overfitting problem.

Adverse neonatal outcome prediction experiments are conducted using the limited data of 121 preterm neonates. Each Electronic Health Record (EHR) consists of 69 input attributes and three primary neonatal outcomes. Specifically, there are two categorical (i.e., severe BPD and PH diagnosis) and one continuous (i.e., discharge weight) adverse neonatal outcome. Ten traditional machine learning algorithms are compared. The F1 score and the Area under the ROC Curve (AUC) are used to evaluate algorithms for the two categorical outcomes. The mean squared error (MSE) is used for the continuous outcome. The empirical results show that each task-specific MTL method outperforms its base single-task model (i.e., NN). To obtain insights into the reasoning behind our model's predictions, we employ  Grad-CAM~\cite{selvaraju2017grad} to estimate the feature importance of 69 input attributes.

Overall, the main contributions of this work are as follows: 1) this work focuses on preterm neonates instead of newborns in general, 2) we propose a novel multi-task learning (MTL) framework for jointly predicting multiple adverse neonatal outcomes using limited annotated data, and 3) we analyze the feature importance for each adverse neonatal outcome to obtain insights into model interpretability.

%% file: figures/data_stat.tex
\begin{figure*}[t]
    \centering
	\begin{minipage}[t]{0.32\linewidth}
    \includegraphics[width=1.0\textwidth]{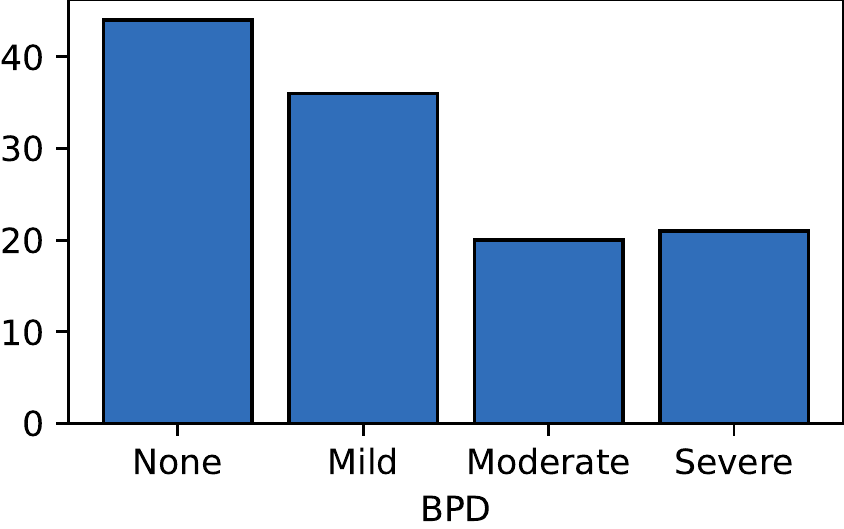}
    \subcaption{Histogram of BPD}
    \label{fig:hist_bpd}
    \end{minipage}
    \hfill
    \begin{minipage}[t]{0.328\linewidth}
    \includegraphics[width=1.0\textwidth]{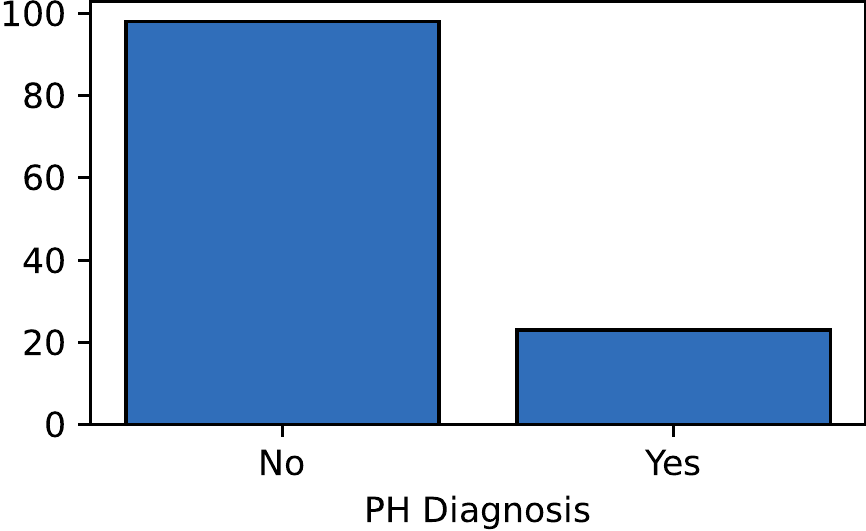}
    \subcaption{Histogram of PH diagnosis}
    \label{fig:hist_phd}
    \end{minipage}
    \hfill
    \begin{minipage}[t]{0.32\linewidth}
    \includegraphics[width=1.0\textwidth]{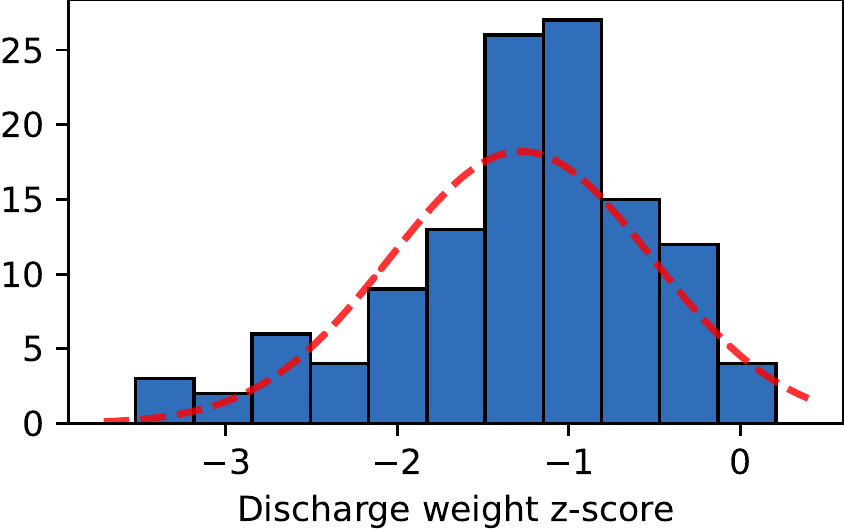}
    \subcaption{Histogram of discharge weight z-score}
    \label{fig:hist_dw}
    \end{minipage}
    \caption{Histograms of three adverse neonatal outcomes, including (a) BPD, (b) PH diagnosis, and (c) discharge weight z-score.}\label{fig:hist_total}
\end{figure*}

%% file: text/relatedwork.tex
\section{Related Work}

{\bf Prediction of neonatal outcomes with machine learning models.} Mboya {\em et al.}~\cite{mboya2020prediction} demonstrate that machine learning models achieve better predictive performance over classical or conventional regression models. Furthermore, Mangold {\em et al.}~\cite{mangold2021machine} conducts a systematic review that collates, critically appraises, and analyzes existing ML models for predicting neonatal outcomes. Unlike Mboya {\em et al.}~\cite{mboya2020prediction} or Mangold {\em et al.}~\cite{mangold2021machine}, Sheikhtaheri {\em et al.}~\cite{sheikhtaheri2021prediction} consider a more practical setting where the experiments are only performed on newborns in neonatal intensive care units (NICUs). Hsu {\em et al.}~\cite{hsu2021machine} leverage ML models (i.e., Random Forest and bagged CART) to estimate the neonatal mortality rate with respiratory failure, which often indicates a higher severity of illness. In contrast to prior research~\cite{mangold2021machine,sheikhtaheri2021prediction,hsu2021machine}, we intend to investigate correlations between different outcomes. Previous works~\cite{hu2022prediction,harutyunyan2019multitask} have shown the effectiveness of a multi-task learning framework for clinical prediction. Therefore, we believe that leveraging these correlations might further improve the ML models. We formulate the prediction of neonatal outcomes as a multi-task learning problem, which considers the potential correlations across multiple outcomes and prevents the ML models from overfitting~\cite{ruder2017overview,zhang2021survey}.

{\bf Multi-task learning.} Multi-task learning (MTL) is a machine learning approach that involves training a single model to perform multiple tasks at the same time~\cite{ruder2017overview, zhang2021survey}. By training a model on multiple tasks simultaneously, it can learn shared features and representations that are useful for all tasks, promoting its generalization ability and making it more robust. MTL has applications in a wide range of fields, including computer vision~\cite{ren2015faster,he2017mask}, natural language processing~\cite{hashimoto2016joint,sogaard2016deep}, and speech synthesis~\cite{wu2015deep}. In the clinical scenario, prior research~\cite{hu2022prediction,harutyunyan2019multitask} has explored the effectiveness of multi-task learning. Hu {\em et al.}~\cite{hu2022prediction} employ an MTL framework to screen commercially available and effective inhibitors against SARS-CoV-2. Harutyunyan {\em et al.}~\cite{harutyunyan2019multitask} build four benchmarks for clinical time series data and propose an MTL framework, which empirically demonstrates that multi-task training acts as a regularizer for almost all tasks. In this work, we first explore the use of MTL in jointly predicting multiple neonatal outcomes for preterm infants.

%% file: text/data.tex
\input{figures/outcome_stat}
\section{Data Collection and Preprocessing}
This section introduces how we collect and preprocess the data. After that, we present a preliminary data analysis on neonatal outcomes.

\subsection{Data Collection}
This study is performed using two combined ELGANs cohorts ($N=184$) from the Prematurity and Respiratory Outcomes Program (PROP) and Prematurity, Respiratory outcomes, Immune System and Microbiome (PRISM) studies across two medical centers (University of Rochester and University at Buffalo). PROP and PRISM were studies performed under Institutional Review Board (IRB) approval (ClinicalTrials.gov: NCT01435187 and ClinicalTrials.gov: NCT01789268). Patients who died before 36 weeks’ Corrected Gestational Age (CGA), have incomplete nutrition or respiratory profiles, or are enrolled in a blinded nutrition study are excluded from analyses. Demographic data is collected to summarize maternal exposures, partner exposures, and other important pregnancy outcomes such as the presence of chorioamnionitis (intrauterine infection). Resuscitation data is collected from interventions necessary after the infant is born in the delivery room. Once admitted to the neonatal intensive care unit (NICU), nutrition and respiratory variables are collected every day. For example, total daily caloric intake is extracted from daily logs and further subdivided by macronutrients (fat, protein, and carbohydrates) coming from parenteral or enteral sources and summarized over the first 28 postnatal days. Respiratory variables from daily flowsheets include the mode of ventilation, pressure/flow, and the fraction of inspired oxygen delivered to the infant.  From this, cumulative oxygen exposure~\cite{dylag2020early}, and counts of conventional and high-frequency ventilator days are obtained and summarized over the first 28 postnatal days. In total, the amount of input attributes is 431.

There are three primary outcomes in this study.  The first is severe BPD, defined as the need for respiratory support at 36 weeks' CGA by the National Institutes of Health Workshop Definition~\cite{fenton2013systematic}.  The second outcome is pulmonary hypertension (yes/no) diagnosed by echocardiography before hospital discharge. The third is the discharged weight z-score, calculated using Fenton premature infant growth curves at birth and 36 weeks' CGA~\cite{higgins2018bronchopulmonary}. 

In summary, we collect Electronic Health Records (EHRs) from 184 preterm neonates. Each neonatal record consists of 431 input attributes and three primary outcomes.

\subsection{Data Preprocessing}
Data preprocessing is essential in the machine learning pipeline, especially for medical data. In our study, we preprocess EHRs from 184 preterm infants, which consists of 431 input attributes for each neonate. Specifically, we will outline the steps involved in our data preprocessing below.

{\bf Data cleaning.}  We first remove duplicate items, which refer to EHR containing identical input attributes. Next, we drop the input attributes with a single value (e.g., care provided in the delivery room). We then remove the input attributes that exhibit an excess of 80\% of missing values (e.g., cord gas 2). Finally, the missing values in the rest of the input attributes are filled in using the Multiple Imputation by Chained Equations (MICE)~\cite{van2011mice}. In particular, the MICE method adopts a series of regression models to iteratively estimate missing values based on available records.

{\bf Data transformation.} The two nominal features (i.e., maternal and partner education) are transformed into numeric variables. We merge the 28-day time-series features (e.g., daily calories, daily fat, daily protein, daily carbohydrates, and daily oxygen exposure) into numerical data by adding up the values for each time-series feature. We proceed by converting categorical variables (e.g., race) into one-hot vectors. Finally, we use z-score normalization for all the remaining features.

In the end, the preprocessed data contain 69 input attributes of 121 neonatal samples. The reduction in the number of samples and the number of input features resulted from the data cleaning and transformation steps, e.g., due to incomplete or inconsistent information.  This preprocessed data is then used for further analysis to train and evaluate machine learning methods. Due to the limited available labeled data in this study, it is critical to investigate techniques such as MTL to improve the data efficiency.

\subsection{Preliminary Data Analysis on Neonatal Outcomes}

{\bf Data distribution.} We present the distributions of three neonatal outcomes in Fig.~\ref{fig:hist_total} (i.e., BPD, PH diagnosis, and discharge weight).  In the case of both categorical variables (i.e., BPD and PH diagnosis), there is a class-imbalanced issue as shown in Fig.~\ref{fig:hist_bpd} and Fig.~\ref{fig:hist_phd}. For BPD, the majority class (BPD: None) accounts for 36\%, while the minority class (BPD: Moderate) only represents 16\%. For pulmonary hypertension diagnosis, 19\% are positive, while 81\% are negative. In addition, Fig.~\ref{fig:hist_dw} suggests that the discharge weight z-score roughly follows a Gaussian distribution.

\input{figures/framework}

{\bf Correlation among adverse neonatal outcomes.} Multi-task learning allows the model to learn shared features relevant to all tasks, which could improve the performance of each individual task. Therefore, it is necessary to conduct an empirical evaluation to determine the relevance of the three neonatal outcomes in our dataset. To this end, we investigate the correlation among all three neonatal outcomes as shown in Fig.~\ref{fig:corr}. 

\begin{itemize}
    \item {\bf BPD and discharge weight.} For each category of BPD, we compute the mean and standard deviation of the discharge weights, respectively, and then adopt these values to plot the normal distribution curve. Fig.~\ref{fig:dw_bpd} shows that BPD and discharge weight are negatively correlated, which means that as the severity of BPD increases, the infants' discharge weights are more likely to decrease.

    \item {\bf PH diagnosis and discharge weight.} Similarly, we show the correlation between PH diagnosis and discharge weight in Fig.~\ref{fig:dw_phd}, which indicates that PH diagnosis and discharge weight are negatively correlated. If an infant is diagnosed with PH, its discharge weight tends to be lighter.

    \item {\bf BPD and PH diagnosis.} We present a 2D histogram in Fig.~\ref{fig:phd_bpd} to visualize the correlation between BPD and PH diagnosis. The data suggest that there is a slightly positive correlation between BPD and a diagnosis of PH, which means that infants who do not develop BPD are less likely to be diagnosed with PH. As the severity of BPD rises to ``moderate'', the likelihood of PH also increases.
\end{itemize}

%% file: figures/outcome_stat.tex
\begin{figure*}[t]
    \centering
	\begin{minipage}[t]{0.32\linewidth}
    \includegraphics[width=1.0\textwidth]{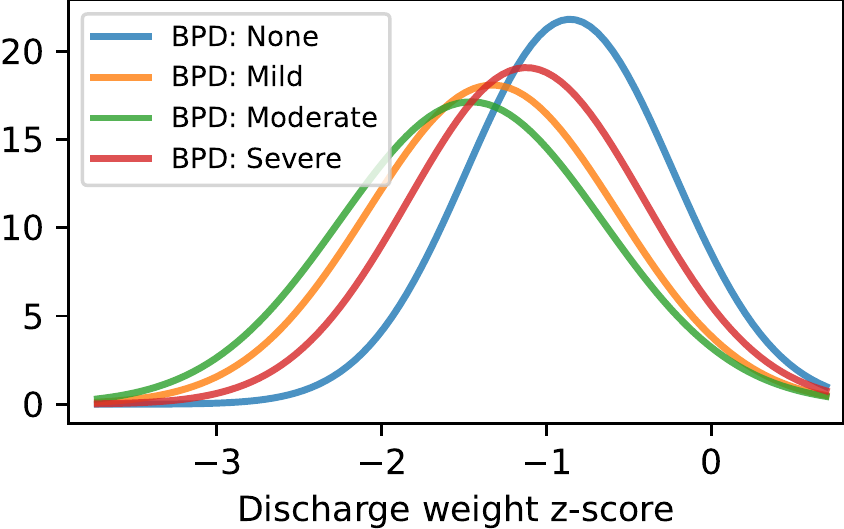}
    \subcaption{BPD vs. Discharge Weight}
    \label{fig:dw_bpd}
    \end{minipage}
    \hfill
    \begin{minipage}[t]{0.32\linewidth}
    \includegraphics[width=1.0\textwidth]{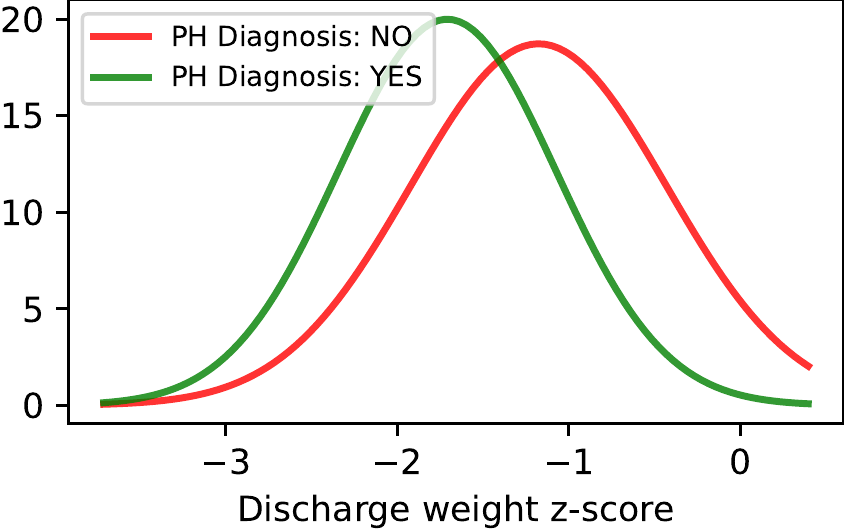}
    \subcaption{PH Diagnosis vs. Discharge Weight}
    \label{fig:dw_phd}
    \end{minipage}
    \hfill
    \begin{minipage}[t]{0.32\linewidth}
    \includegraphics[width=1.0\textwidth]{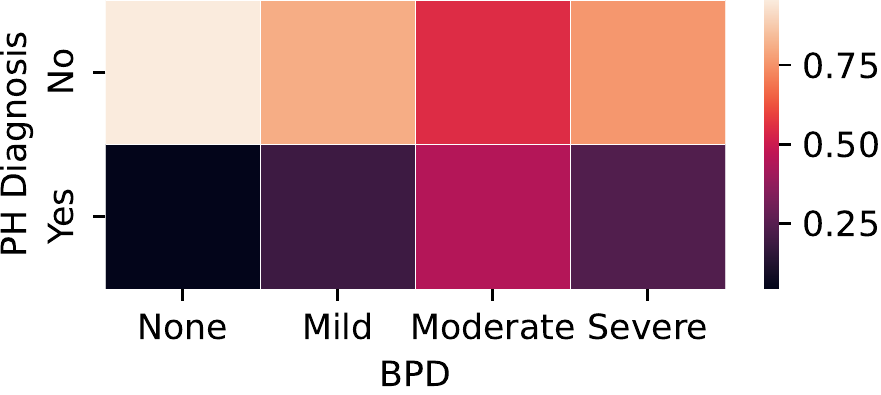}
    \subcaption{PH Diagnosis vs. BPD}
    \label{fig:phd_bpd}
    \end{minipage}
    \caption{Correlation among different adverse neonatal outcomes, including (a) BPD vs. discharge weight, (b) PH diagnosis vs. discharge weight, and (c) PH diagnosis vs. discharge weight.}\label{fig:corr}
    \vspace{-2mm}
\end{figure*}

%% file: figures/framework.tex
\begin{figure*}[t]
    \centering
	\begin{minipage}[t]{0.39\linewidth}
    \includegraphics[width=1.0\textwidth]{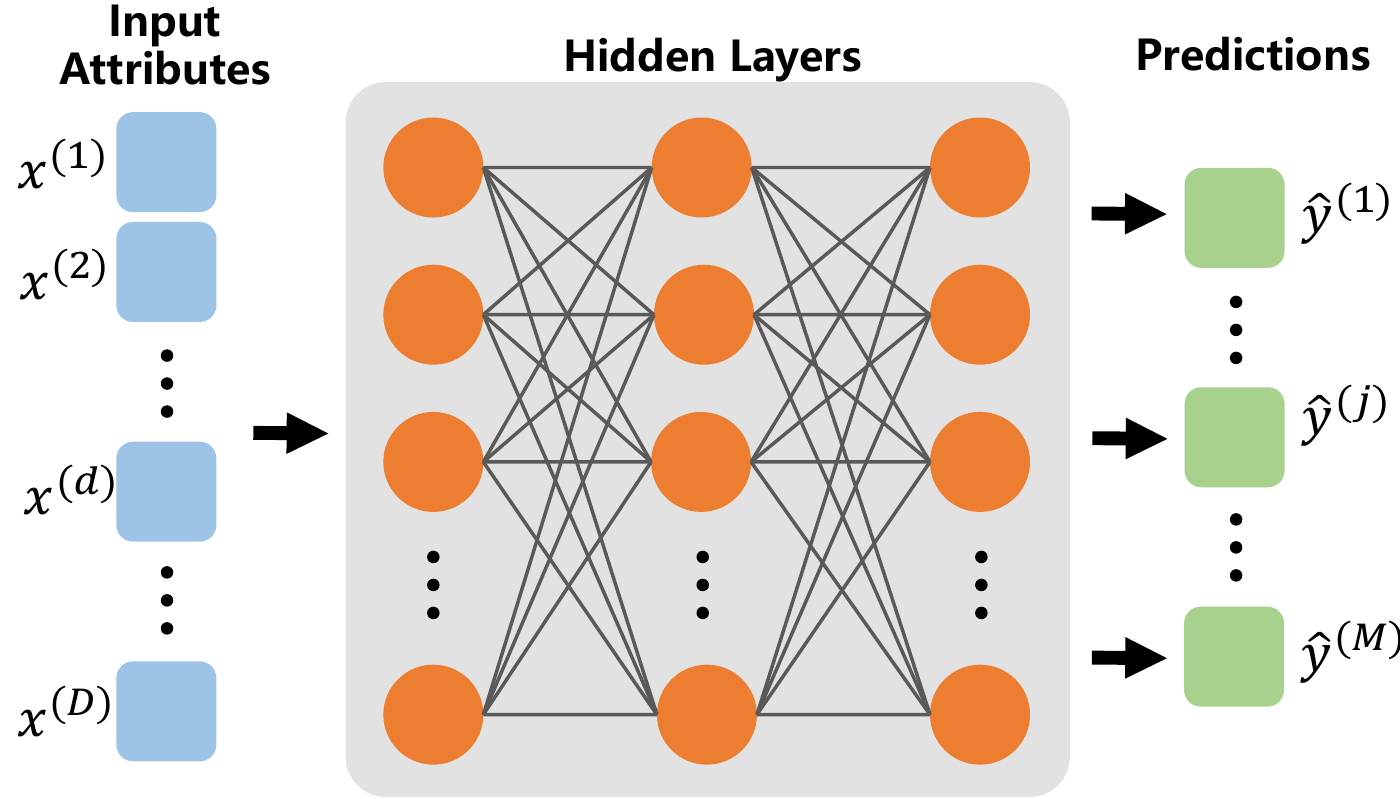}
    \subcaption{Single-task learning baseline}
    \label{fig:single}
    \end{minipage}
    \hfill
    \begin{minipage}[t]{0.58\linewidth}
    \hfill
    \includegraphics[width=1.0\textwidth]{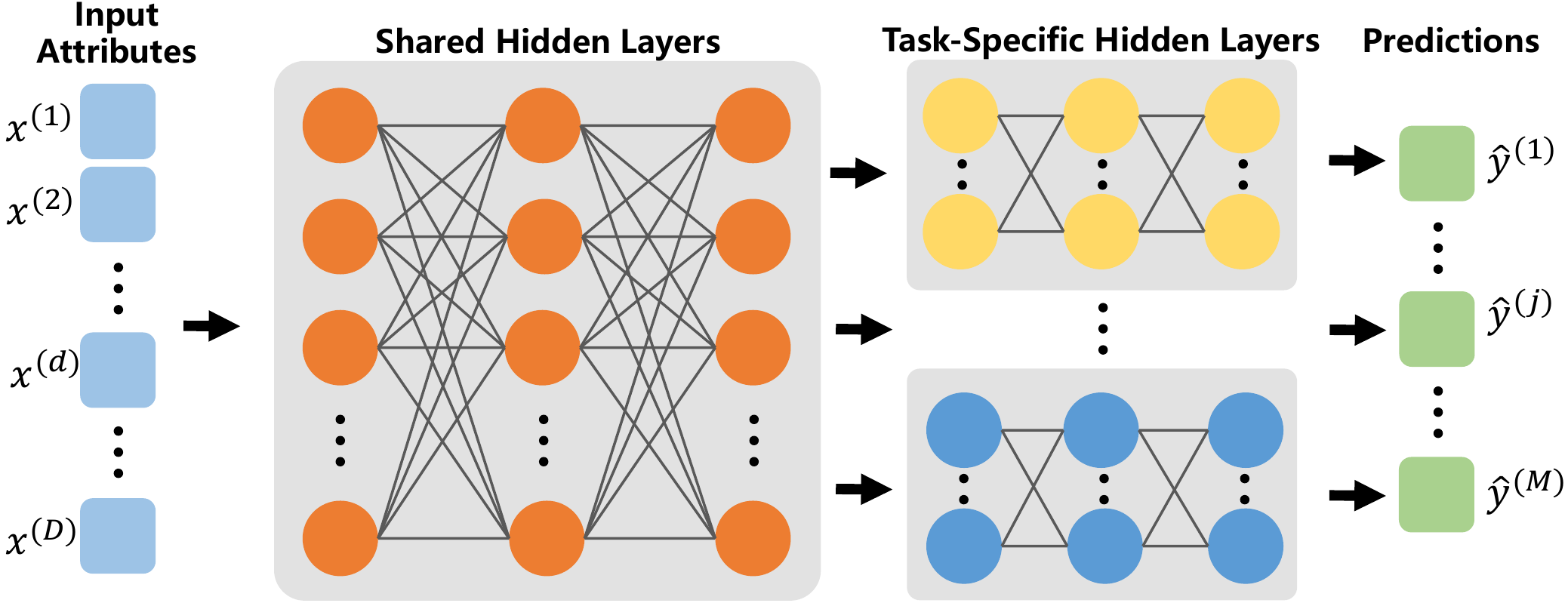}
    \subcaption{Multi-task learning framework}
    \label{fig:multi}
    \end{minipage}
    \caption{Comparison between (a)  single-task learning baseline and (b) multi-task learning framework. In the multi-task learning framework, input features $\{x^{(d)}\}_{d=1}^{D}$ are first processed by shared hidden layers to produce a latent feature representation. This latent representation is then passed through task-specific hidden layers to generate multiple predictions $\{\hat{\by}^{(j)}\}_{j=1}^{M}$.}
     \vspace{-4mm}
\end{figure*}

%% file: text/method.tex
\section{Methodology}

In this section, we first formulate the problem. We then describe the single-task learning baseline (i.e., Neural Network) for predicting each adverse neonatal outcome. Finally, we present the proposed multi-task learning framework for multiple outcome predictions.

\subsection{Problem Formulation}
We formulate the prediction of adverse neonatal outcomes as a multi-task learning problem. The input feature ${\bx}_i \in \mathbb{R}^{1\times D}$ consists of information about the infant, its parents,  delivery room, nutrition summary, etc. $D$ denotes the number of input features, and $x_i$ denotes the $i\text{-th}$ sample in the dataset. $\by = \{\by^{(j)}\}_{j=1}^{M}$ refers to the ground truth of different adverse neonatal outcomes, where $M$ indicates the total number of adverse neonatal outcomes and $\by^{(j)}$ is the $j\text{-th}$ outcome. For the MTL framework, we define the shared hidden layers as $f(\cdot; \theta)$ transferring input features $\bx$ into latent features $f(\bx; \theta)$, where $\theta$ denotes the learnable parameters. Following the shared hidden layers, $M$ task-specific branches $\{g^{(j)}(\cdot; \theta^{(j)})\}_{j=1}^{M}$ transfer the shared latent features into the final predictions $\hat{\by}_i$ for different neonatal outcomes:
\begin{equation}
    \hat{\by}_i=\{g^{(j)}(f(\bx_i; \theta); \theta^{(j)})\}_{j=1}^{M},
\end{equation}
where $\hat{\by}_i$ denotes the prediction of the $i$-th sample with $M$ outcomes, and $\theta^{(j)}$ represents the parameters of the $j$-th task-specific branch.

\subsection{Single-Task Learning Baseline}
As shown in Fig.~\ref{fig:single}, we employ a Neural Network with multiple hidden layers as our backbone. In particular, the neural network $\text{NN}(\cdot; \theta)$ converts an input sample $\bx_i$ with $D$ input attributes into a prediction $\hat{\by}_i^{(j)}$ for the $j$-th neonatal outcome.

For the classification task, we obtain the prediction scores $\hat{\by}^{(j)}$ using a softmax function. After that, we use the cross-entropy loss to optimize the classifier:
 \vspace{-1mm}
\begin{equation}
    \label{eq:cls}\mathcal{L}_\mathtt{cls}^{(j)} = -\frac{1}{N} \sum^{N}_{i=1} \by_i^{(j)} \cdot \log(\hat{\by}_i^{(j)}),
\end{equation}
 \vspace{-2mm}

where $\by_i^{(j)}$ denotes the ground truth of the $i\text{-th}$ training sample for the $j\text{-th}$ task, and $N$ denotes the batch size.

For the regression task, we use the mean squared error (MSE) loss to optimize the regression model:
 \vspace{-1mm}
\begin{equation}
    \label{eq:reg}
    \mathcal{L}_\mathtt{reg}^{(j)} = \frac{1}{N}\sum^{N}_{i=1} \Vert \by_i^{(j)} - \hat{\by}_i^{(j)} \Vert^2_2.
\end{equation}
 \vspace{-2mm}

\input{tables/mtl_exp}

\subsection{Multi-Task Learning Framework}

Although previous ML-based works have shown the effectiveness of the single-task learning framework, such a framework might ignore the potential correlations between different outcomes, leading to suboptimal results. Moreover, single-task learning models are more likely to suffer from overfitting issues, especially in cases with limited training data. 
As shown in Fig.~\ref{fig:multi}, we propose a novel multi-task learning framework to leverage the potential correlation between various adverse neonatal outcomes and avoid the overfitting problem. Technically, the proposed multi-task learning framework consists of shared hidden layers and multiple task-specific branches. Each task-specific branch contains several hidden layers and a prediction layer. The overall objective $\mathcal{L}_\mathtt{mtl}$ of the MTL framework is computed by combining the weighted losses of multiple tasks:
\begin{equation}
    \mathcal{L}_\mathtt{mtl} = \sum^{M}_{j=1} \lambda^{(j)} \mathcal{L}_\mathtt{stl}^{(j)},
\end{equation}
where $\mathcal{L}_\mathtt{stl}^{(j)}$ denotes the $j$-th task-specific objective with the loss weight $\lambda^{(j)}$, and $M$ is the number of tasks.

Overall, the proposed MTL framework can exploit the correlations among neonatal outcomes since by using multi-task learning, the framework is better equipped to capture the general patterns that are relevant to all tasks. These patterns would promote the generalization ability and prevent ML-based methods from overfitting~\cite{zhang2021survey,ruder2017overview}. Meanwhile, the MTL framework can enhance the performance of a specific primary task by using related tasks as auxiliary tasks~\cite{liebel2018auxiliary}, which provide auxiliary information for the primary task, resulting in a more efficient and effective learning process.

%% file: tables/mtl_exp.tex
\begin{table*}[t]
\caption{Overall performance on prediction of adverse neonatal outcomes. The BPD and PH Diagnosis refer to bronchopulmonary dysplasia and pulmonary hypertension diagnosis, respectively. The best score for each task is highlighted in bold. The improvements in task-specific MTL methods are shown in red. The results of BPD and PH Diagnosis are in \%.}
\begin{adjustbox}{width=1\textwidth}\setlength\tabcolsep{2pt}
\begin{tabular}{lccccc}
\toprule[1pt]
\multicolumn{1}{c}{\multirow{2}{*}{Methods}} & \multicolumn{2}{c}{\textbf{BPD} (Task 1)}                                 & \multicolumn{2}{c}{\textbf{PH Diagnosis} (Task 2)} & \textbf{Discharge Weight} (Task 3) \\ \cmidrule(r){2-3} \cmidrule(r){4-5} \cmidrule(r){6-6}
\multicolumn{1}{c}{}                         & F1 $\uparrow$    & AUC $\uparrow$ & F1 $\uparrow$              & AUC $\uparrow$             & MSE $\downarrow$                       \\ \midrule[1pt] 
Naive Bayes                       & 22.5 & 60.4 & 33.4           & 59.7           & -                         \\
Logistic Regression~\cite{cox1958regression}                           & 40.2 & 69.0          & 28.0           & 66.4           & -                         \\
Random Forest~\cite{ho1995random}                                 & 42.4 & 81.2 & 44.3           & 75.3  & 0.266            \\
Decision Tree~\cite{wu2008top}                                 & 46.9 & 76.9       & 35.1           & 55.2           & 0.353                     \\
SVM/SVR~\cite{cortes1995support}                                           & 39.2 & 71.0    & 25.9           & 43.2           & 0.293                     \\
XGBoost~\cite{Chen:2016:XST:2939672.2939785}                                       & 31.8          & 72.4  & 39.8           & \textbf{77.2}           & 0.296                     \\
Ridge Regression~\cite{hoerl1970ridge}                              & -              & -              & -               & -               & 0.365                     \\
Lasso Regression~\cite{tibshirani1996regression}                              & -              & -              & -               & -               & 0.270                     \\
ElasticNet~\cite{zou2005regularization}                                    & -              & -              & -               & -               & 0.267                     \\ \midrule[1pt] 
Neural Network~\cite{mcculloch1943logical} (base model)                   &  43.1         &     81.2     & 32.3           & 75.1           & 0.279                     \\
MLT (tuned for BPD)  &    {\hspace{6.8mm} \textbf{48.0} \color{red}{($\uparrow$4.9)}}   &  {\hspace{6.8mm} \textbf{83.2}  \color{red}{($\uparrow$2.0)}}   & 39.7          & 68.2           & 0.288                     \\
MLT (tuned for PH Diagnosis)                                  & 35.8          & 65.7          & {\hspace{6.85mm} \textbf{46.6} \color{red}{($\uparrow$2.3)}} & {\hspace{6.85mm} 76.5 \color{red}{($\uparrow$1.4)}}  & 0.330                     \\
MLT (tuned for Discharge Weight)                                & 43.2          & 79.8         &     42.0       & 71.8           & {\hspace{10mm} \textbf{0.257} \color{red}{($\downarrow$0.022)}}                    \\ \bottomrule[1pt] 
\end{tabular}
\end{adjustbox}\label{tab:overall}
\end{table*}

%% file: text/experiment.tex
\section{Experiment}

\subsection{Experiment Setup}
{\bf Dataset.} Adverse neonatal outcome prediction experiments are conducted using the data of 121 premature infants. Each sample has 69 input features and three adverse neonatal outcomes. In particular, two adverse neonatal outcomes are categorical and the other is numerical. We use $5$-fold cross-validation to obtain a more reliable estimate of the machine learning models' performance.

{\bf Evaluation metric.}  We use the F1 score and Area under the ROC Curve (AUC) of the positive samples to evaluate the performance of the BPD and PH diagnosis classification tasks. We use the mean squared error (MSE)  to estimate the discharged weight prediction, which is a regression task.

\subsection{Baseline Methods}
Extensive experiments have been conducted with various ML techniques on our dataset. We list all ML methods for classification and regression tasks below.

{\bf Classification task.} Six traditional machine learning methods are included in our experiments, including Naive Bayes classifier, Logistic Regression~\cite{cox1958regression}, Random Forest~\cite{ho1995random}, Decision Tree~\cite{wu2008top}, Support Vector Machine (SVM)~\cite{cortes1995support}, and XGBoost~\cite{Chen:2016:XST:2939672.2939785}.

{\bf Regression task.} We implement seven machine learning models to predict the discharged weights of infants. These models are Ridge Regression~\cite{hoerl1970ridge}, Lasso Regression~\cite{tibshirani1996regression}, ElasticNet~\cite{zou2005regularization}, Random Forest~\cite{ho1995random}, Decision Tree~\cite{wu2008top}, Support Vector Regression (SVR)~\cite{cortes1995support}, and XGBoost~\cite{Chen:2016:XST:2939672.2939785}.

\subsection{Implementation Details}
\textbf{Architecture.} Our models are implemented in PyTorch. For each single-task model, we select the optimal number of hidden layers in the range of \{1,2,3,4,5\}. Additionally, we choose the number of neurons per hidden layer from the set of \{128, 256, 512, 1024\}. The hyperparameters of the multi-task model are searched in terms of \{1,2,3,4\}-layer shared hidden layers with \{64, 128, 256, 512\} neurons and \{1,2,3\}-layer task-specific hidden layers with \{64, 128, 256\} neurons. All these hyperparameters are tuned corresponding to the performance on the $5$-fold cross-validation.

\textbf{Optimization.} We optimize the model through an Adam~\cite{kingma2014adam} optimizer with an initial learning rate of $\{5\times10^{-3}, 1\times10^{-2}, 2\times10^{-2}\}$. The batch size is 64, and weight decay is $\{10^{-1}, 10^{-2}, 10^{-3}\}$. The learning rate follows a cosine decay schedule~\cite{loshchilov2016sgdr}. We train all baseline models for \{20, 50, 100\} epochs on a 1080Ti GPU.

\subsection{Comparison with Other Traditional Machine Learning Techniques}
Table~\ref{tab:overall} summarizes the quantitative results on three tasks about the adverse neonatal outcomes. We compare our proposed multi-task learning framework with several traditional machine learning frameworks. 
Overall, the results across different tasks show that our proposed MTL framework outperforms other traditional machine learning methods. Note that Neural Network (NN) is a degraded version of our proposed multi-task learning framework. In other words, comparisons between NN and our MTL methods can clearly ablate the impact of multi-task learning. The comparison between the NN and different task-specific MTL methods suggests that MTL methods can significantly outperform their base model, as shown in Table~\ref{tab:overall}. The improvements indicate that multi-task learning can boost the performance of the primary task by leveraging the other relevant auxiliary tasks, which provide additional information for learning the primary task. Therefore, data efficiency highlights the necessity of exploring MTL techniques, especially for small datasets.  

\subsection{Feature Importance Analysis}
Feature importance analysis provides insight into the relationships between input features and the predicted outcomes. Gradient-weighted Class Activation Mapping (Grad-CAM)~\cite{selvaraju2017grad} generates a heatmap highlighting the regions of an input image most influential for the network's prediction. Grad-CAM has gained widespread usage in the field of computer vision. We adapt the Grad-CAM method to the neonatal health area. In particular, we employ the Grad-CAM to analyze the feature importance in predicting adverse neonatal outcomes.

{\bf Severe BPD.} We list the top 10 most important features for predicting severe BPD:
\begin{itemize}
    \item Total ventilator days ({\bf the most influential})
    \item Oxygen AUC 28 days
    \item Summary total oxygen exposure
    \item Pulmonary deterioration oxygen adjusted
    \item Total conventional ventilator days
    \item Total high-frequency ventilator days
    \item Pulmonary deterioration AUC
    \item Oxygen severity index 28 days
    \item Maternal Smoking
    \item Delivery room hyperthermia
\end{itemize}
Clinicians believe severe BPD should be most correlated with indicators related to the respiratory system, such as ventilator usage~\cite{gibbs2020ventilation,gupta2009ventilatory}, oxygen exposure~\cite{nesterenko2008exposure}, pulmonary deterioration~\cite{davidson2017bronchopulmonary}, maternal smoking~\cite{aschner2017can,morrow2017antenatal}, and so on. The feature importance computed by Grad-CAM largely agrees with the clinicians. Specifically, \textbf{the usage of the ventilator, exposure to oxygen, and pulmonary deterioration are the most critical factors in predicting severe BPD}, as expected. In addition, it is found that maternal smoking has a significant impact on the severity of BPD. 

{\bf PH Diagnosis.} Similarly, the top 10 most important features of pulmonary hypertension diagnosis are shown as follows:
\begin{itemize}
    \item Birth weight z-score ({\bf the most influential})
    \item respiratory failure reintubations provided
    \item Delivery room - CPAP provided
    \item Birth weight less than 10th percentile (Intrauterine growth restriction)
    \item Delivery room - intubated
    \item Delivery mode 1
    \item Delivery room hyperthermia
    \item Delivery room - resuscitation provided
    \item Delivery room - chest compressions
    \item Summary total protein
\end{itemize}
Clinicians suggest that PH diagnosis is closely associated with birth weight~\cite{bhat2012prospective}, lung conditions~\cite{nathan2019pulmonary}, delivery mode~\cite{wilson2011persistent}, etc. The results of feature importance are \textit{aligned} with the expectations of clinicians, as \textbf{birth weight and respiratory failure reintubations are the primary factors of the PH diagnosis}. Moreover, the results show the conditions of the delivery room are critical factors in PH diagnosis. Furthermore, the nutrition inputs (e.g., protein) influence the PH diagnosis.

{\bf Discharge weight.} The following are the top 10 most important features, in order, for predicting the discharge weight:
\begin{itemize}
    \item Birth Weight z-score ({\bf the most influential})
    \item Infant gestational age
    \item Infant birth weight
    \item Delivery room - intubated
    \item Delivery room hyperthermia
    \item Delivery room - CPAP provided
    \item Delivery room - chest compressions
    \item Delivery mode 1
    \item Maternal smoking
    \item Maternal asthma diagnosis
\end{itemize}
The results show that {\bf discharge weight is most related to infant birth weight and gestational age.} In addition, the delivery mode and conditions of the delivery room have significant impacts on discharge weight.

%% file: text/conclusion.tex
\section{Conclusion and Future Work}
In this paper, we investigate multi-task learning for predicting multiple adverse neonatal outcomes. In contrast to previous ML-based methods that formulate this problem as a single-task problem, our proposed multi-task learning framework jointly learns several relevant tasks to capture the shared general patterns across various tasks. To implement our idea, we propose an MTL framework with several shared hidden layers followed by task-specific branches, which integrates shared general patterns into the ML models, while remains task-specific information for different adverse neonatal outcomes. Extensive experiments conducted on three primary neonatal outcomes demonstrate the effectiveness of the proposed multi-task learning method trained on a limited dataset. We hope our study will draw the community’s attention to the diagnosis of multiple neonatal outcomes for premature infants.
For future work, we plan to investigate the domain adaptation capability of the MTL framework by evaluating its performance on additional test samples obtained from other medical facilities.

\section*{Acknowledgments} 
The study was supported in part by K08HL155491 (AM Dylag), The Prematurity and Respiratory Outcomes Program (PROP): National Institutes of Health, NHLBI and NICHD through U01 HL101813 to University of Rochester and University at Buffalo (GS Pryhuber). NIH/NIAID HHSN272201200005C University of Rochester Respiratory Pathogens Research Center (URRPRC).